\documentclass[fleqn,10pt,twocolumn]{SWARM17}

\title{One-shot path planning for multi-agent systems \\using fully convolutional neural network}

\author{Tomas Kulvicius${}^{1}$, Sebastian Herzog${}^{1*}$, Timo L\"uddecke${}^{1*}$, \\Minija Tamosiunaite${}^{1,2}$ and Florentin W\"org\"otter${}^{1\dagger}$}
\speaker{Florentin W\"org\"otter}

\affils{${}^{1}$Third Institute of Physics, University of G\"ottingen, G\"ottingen, Germany\\
${}^{2}$Faculty of Informatics, Vytautas Magnus University, Kaunas, Lithuania\\
(E-mail: tomas.kulvicius@uni-goettingen.de)\\
}

\abstract{%
Path planning plays a crucial role in robot action execution, since a path or a motion trajectory for a particular action has to be defined first before the action can be executed. Most of the current approaches are iterative methods where the trajectory is generated iteratively by predicting the next state based on the current state. Moreover, in case of multi-agent systems, paths are planned for each agent separately. In contrast to that, we propose a novel method by utilising fully convolutional neural network, which allows generation of complete paths, even for more than one agent, in one-shot, i.e., with a single prediction step. We demonstrate that our method is able to successfully generate optimal or close to optimal paths in more than 98\% of the cases for single path predictions. Moreover, we show that although the network has never been trained on multi-path planning it is also able to generate optimal or close to optimal paths in 85.7\% and 65.4\% of the cases when generating two and three paths, respectively.}


\begin{document}

\maketitle


\section{Introduction}

Motion (path) planning is one of the fundamental issues in the development of truly autonomous robots, let it be a mobile robot or a robotic-manipulator. In robotics, motion planning is defined as the problem of finding a temporal sequence of valid states (e.g., robot positions or configurations), which brings a robot (-arm) from a start- to a goal-position (configuration) given some constraints (e.g., obstacles) \cite{Latombe2012}. In this work, we specifically address the issue of multiple path planning for multi-agent systems.

Most common classical approaches for path planning are the Dijkstra algorithm \cite{Dijkstra1959}, the A* search \cite{Hart1968}, and its variants (e.g., see \cite{Korf1985,Koenig2004,Sun2008}). Dijkstra and A* algorithms perform well on grid-based representations and provide the optimal solution (i.e., shortest path). However, they do not scale well with increased dimensions and path lengths.

Another class of common approaches, are sampling based methods such as rapidly-exploring random tree algorithm (RRT, \cite{LaValle1998}) and it is variants (e.g., see \cite{Karaman2011,Islam2012,Gammell2014}). While RRTs are more suitable for continuous spaces, they do not perform so well on grids as compared to Dijkstra or A* path search \cite{Knispel2013,Bency2019}, i.e., paths are not necessarily optimal. Also these methods are computationally more expensive and require parameter tuning to obtain optimal performance, whereas Dijsktra and A* are parameter-free methods.

Some other path planning approaches are based on bio-inspired neural networks \cite{Glasius1995,Glasius1996,Bin2004,Yang2001,Ni2017,Rueckert2016}. The neurons in the network represent specific locations in the environment similar to place cells found in hippocampus \cite{OKeefe1971}. The optimal path is found by activating the target neuron and propagating activity to the neighbouring cells until the source neuron is reached. The path is reconstructed by following activity gradient of the network from the source to the target. Conceptually, these methods are similar to the Dijkstra algorithm and will find the optimal solution, however, these approaches require several iterations until the solution is found.

Recently, several path planning methods have been proposed using deep learning approaches such as deep multi-layer perceptrons (DMLP, \cite{Qureshi2018}), long short-term memory (LSTM) networks \cite{Bency2019}, and deep reinforcement learning (deep-RL) approaches \cite{Tai2017,Panov2018}. All these methods generate paths iteratively by predicting the next state based on the environment configuration, the current state and the target position until the target is reached. Thus, the network has to be exploited many times until a complete path can be constructed.

In case of path planning for multi-agent systems, usually, decentralised approaches are used (e.g., see \cite{Wang2011,Desaraju2012,Chen2017,Everett2018,Long2017}), where path search is performed for each agent separately. Thus, the computation time scales with the number of the searched paths (i.e, the number of agents). Different from the afore mentioned approaches, in this study we present a method based on a fully convolutional network (FCN), which allows multi-path planning in one-shot, i.e., complete multiple paths can be generated by our network with a single prediction iteration.

The structure of this manuscript is as follows. In the following section we explain our approach (section 2), then we provide an evaluation of our method where we first analyse the general performance of the proposed network and afterwards we analyze path planning of multiple paths (section 3). Finally, in section 4, we discuss our results and relate our approach to similar methods.

\section{Methods}

\subsection{Overview}

\begin{figure*}[htb]
\begin{center}
\includegraphics[width=0.9\linewidth]{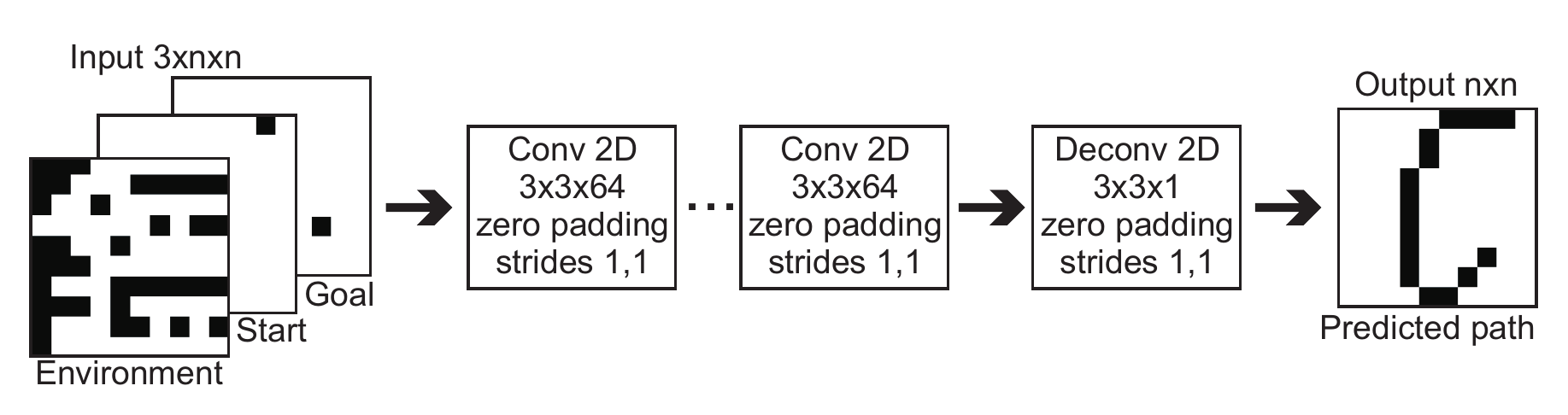}
\caption{\label{net} Proposed network architecture. We used a fully convolutional neural network (FCN) with 20 convolutional layers and one deconvolutional layer (transpose convolution). We used ReLU activation units in the convolutional layers and a sigmoid activation unit in the deconvolutional layer (the last layer). We used a batch normalisation layer after each convolutional layer and in addition a dropout (10\%) layer  after the deconvolutional layer (not shown).  For more details please refer to the main text.}
\end{center}
\end{figure*}

The task is to predict an optimal path (or several paths) for an unknown environment with obstacles (not used in the training set), given any start- and end-point (goal). In this study, we only analysed 2D environments but extension of the presented method to 3D cases is also possible and is straightforward.

Each environment is described by a grid (binary image) of size $n \times n$ where free spaces are marked in white color and not traversable spaces (obstacles) are marked in black. The start- an end-points are  represented by separate binary images where the start- (or end-point) is marked black and all other grid cells are marked white (see input in Fig.~\ref{net}).

After training, the model is able to predict a collision-free path from the given start- and end-point which is represented as a binary image where the path is marked again in black (see output in Fig.~\ref{net}). The actual trajectory, i.e. the sequence of locations that the agent needs to traverse, is then constructed by tracking the black cells using forward- (from the start-point) and backward- (from the end-point) search until the path-segments meet or cross each other.

In the following section we provide a detailed description for each of the afore discussed steps.

\subsection{Data}

\textbf{Input definition:} As discussed above, each environment is described by a $n \times n$ grid map. We define the grid map as a binary image $I^e$, where we set $I^e_{i,j} = 0$, if the grid cell $\{i,j\}$ is free (no obstacles), otherwise we set $I^e_{i,j} = 1$ ($i=1\dots n, j=1\dots n$), i.e., if it contains obstacles. Similarly, we also define maps for start- and end-points: we set $I^{s/g}_{i,j} = 1$ at the start-/end-point, and we set $I^{s/g}_{i,j} = 0$ anywhere else. Here $I^s$ and $I^g$ denote the start map and the goal map, respectively. Thus, we obtain a 3D input of size $3 \times n \times n$.

We have also analysed the network's capability to generalise and predict multiple-paths at once from several start-positions to the same goal position. In these cases several grid cells of $I^s_{i,j}$ were set to $1$ to mark start-positions of the agents.

\textbf{Output definition:} As in case of the input maps, we used a binary map to define the output map $O$, where we set $O_{i,j}=1$, if the found path was traversing this grid cell $\{i,j\}$, and we set $O_{i,j}=0$ everywhere else. Thus, we obtain a 2D output of size $n \times n$. In our study, we used the A* grid search algorithm to find optimal solutions (ground truth paths for training), where eight movement directions (vertical, horizontal and diagonal) were allowed. For vertical and horizontal moves we used a cost of $1$ and for diagonal moves we had cost of $\sqrt{2}$. We used the Euclidean distance from the current grid cell to the goal cell to calculate the cost.

\textbf{Data generation:} The input maps for training and testing were generated randomly in the following way. Each grid cell was set to $1$ (an obstacle) with a probability $p_o=0.6$ or was set to $0$ (a free space) with a probability $p_f=1-p_o$. Note that two diagonal configurations of obstacles were not allowed: $\{I^e_{i,j}=0,I^e_{i,j+1}=1,I^e_{i+1,j}=1,I^e_{i+1,j+1}=0\}$ and $\{I^e_{i,j}=1,I^e_{i,j+1}=0,I^e_{i+1,j}=0,I^e_{i+1,j+1}=1\}$, as this would have allowed the A* algorithm to generate paths going between two diagonally arranged obstacles that touch each other only by a corner, which in real scenarios, however, cannot be traversed.

\begin{figure*}[htb]
\begin{center}
\includegraphics[width=0.99\linewidth]{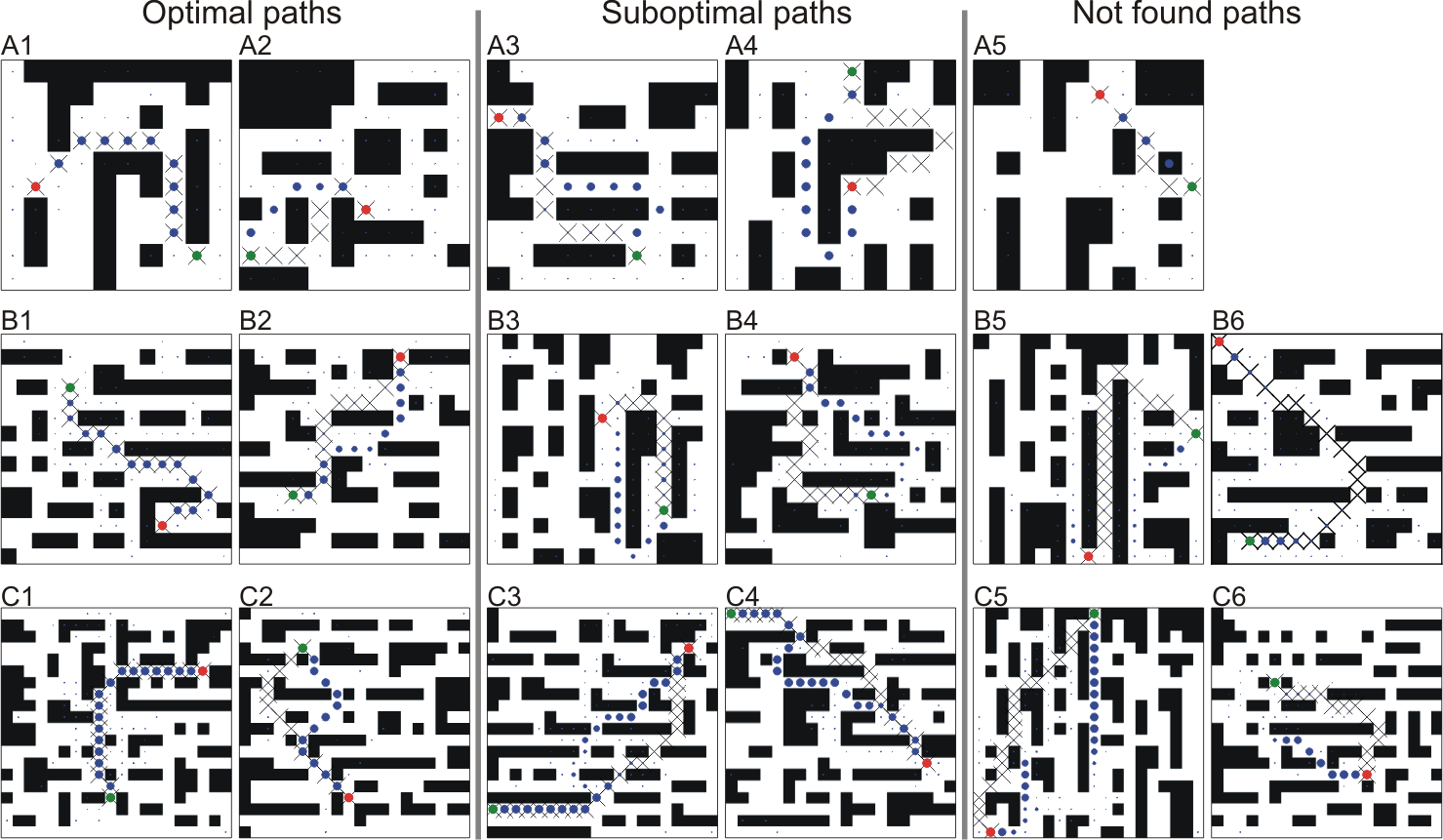}
\caption{\label{exampl_single} Examples of single path predictions on unknown environments for different grid sizes (trained and tested on the same grid size): \textbf{A)} $10 \times 10$, \textbf{B)} $15 \times 15$ and \textbf{C)} $20 \times 20$. The first two columns show optimal (shortest) paths, the middle two columns suboptimal paths and the last two columns not found paths. Note that for $10 \times 10$  (A) we had only one sample of a not found path. Crosses denote the A* solution where blue dots denote the predicted path using our FCN. Size of the dots corresponds to small (close to zero) and large values (close to one) of network outputs. Green and red dots correspond to the start- and end-point, respectively.}
\end{center}
\end{figure*}

We generated two datasets: 1) for learning and testing predictions of single paths and 2) for testing predictions of multiple paths (up to three start-positions). Note that we did not train our network on multiple paths. For the first case, we generated environments of three different sizes: $10 \times 10$, $15 \times 15$ and  $20 \times 20$ (for some examples of environments see Fig.~\ref{exampl_single}). For each case, we generated 30000 samples with different obstacle configurations and different random start- and end-points. Note that the minimum Euclidean distance between start- and end-points was $5$ to avoid very short paths and exclude trivial cases. 28000 samples were used for training and 2000 samples for testing. For the second case, we generated 1000 samples of size $15 \times 15$ with different obstacle configurations but fixed start- and end-positions. Start- and end-points were located at positions $\{1,1\}$, $\{1,n\}$ and $\{n,1\}$ (in the corners) where the goal was at the position $\{8,8\}$ (in the middle). Some examples of environments are shown in Fig.~\ref{exampl_multi}.

\subsection{Proposed Network}

\textbf{Network architecture:} We used a fully convolutional network as shown in Fig.~\ref{net}. The input layer is a 3D input consisting of three 2D binary images of size $n \times n$ and the output is a single 2D binary image of size $n \times n$ as described above. We used 20 identical convolutional layers with 64 filters of size $3 \times 3$ and with stride 1. After each convolutional layer we used a batch normalisation layer (not shown). After the convolutional layers, we added one deconvolutional layer with one filter of size $3 \times 3$. This was followed by a batch normalisation layer and, in addition, a dropout (10\%) layer. In all convolutional layers we used ReLU activation units, whereas in the deconvolutional layer a sigmoid activation unit was used. We define the network output as $\hat{O}$, where $\hat{O_{i,j}}$ can obtain values between 0 and 1. In all layers we used zero padding to keep the same dimension and prevent information loss. Note that we also tried a network architecture with non-zero padding, but performance was worse.

\textbf{Learning procedure:} We used the mean squared error (MSE) between the network output $\hat{O}$ and the ground truth solution $O$ as loss. The ADAM optimiser with default learning parameters and a batch size of 64 samples was used for training the network. Early validation stopping was used if the accuracy on validation set did not increase within the last 10 epochs to prevent over-fitting. We used 28000 samples to train the network (26000 samples for training and 2000 samples for validation) and 2000 samples for testing the network's performance on unseen samples. We trained the network five times and selected the best model with the highest accuracy on the validation set.

For network implementation we used Tensorflow and Keras API\footnote{The source code will be available online after publication of this manuscript.}, where the A* algorithm was implemented in Python 3.6.7. We used a PC with Intel Xeon Silver 4114 (2.2GHz) CPU and NVIDIA GTX 1080 (11GB) GPU.

\subsection{Path reconstruction \label{sec:path_rec}}

As explained above, the output of the network is a value map (grid) of size $n \times n$, which contains values between zeros and one, which can be treated as certainty of the grid-cell to be a part of the path. The output map as such does not yet provide the temporal sequence of points, which would lead from the start-point to the end-point (goal). Thus, the path has to be reconstructed from the prediction-map. In our case, we simply did this by using bidirectional search, which is described as follows. Let us denote the forward and the backward path as a temporal sequence of points on the grid with $\{x^f_t,y^f_t\}$ and $\{x^b_t,y^b_t\}$, respectively. We also annotate the start-point as $\{x_s,y_s\}$ and the end-point (goal) as $\{x_g,y_g\}$. Initially we set $\{x^f_1,y^f_1\}=\{x_s,y_s\}$ and $\{x^b_1,y^b_1\}=\{x_g,y_g\}$. Given the current position $\{x_t,y_t\}$ of the forward/backward path, the next position of the forward/backward path is obtained by choosing the grid cell $\{i,j\}$ in the nearest neighbourhood of the current position (we used eight nearest neighbours) with the maximum value of the network output $\hat{O}$:
\begin{eqnarray}
\{x_{t+1},y_{t+1}\} = \arg\max_{i,j} \hat{O}_{i,j}\,\{i \neq x_{t}, j \neq y_{t}, \\
\nonumber i\in [x_t-1,x_t+1],  j\in [y_t-1,y_t+1] \}.
\end{eqnarray}
After this step, we set $\hat{O}_{i,j}=0$. We continue constructing forward and backward paths until one of the three conditions is met:
\begin{enumerate}
\item End-point is reached, $\{x^f_t,y^f_t\} = \{x_g,y_g\}$;
\item Start-point is reached, $\{x^b_t,y^b_t\} = \{x_s,y_g\}$;
\item The paths cross each other, $\{x^f_t,y^f_t\} = \{x^b_k,y^b_k\}$ or $\{x^b_t,y^b_t\} = \{x^f_k,y^f_k\}$, where $1 \leq k \leq t$.
\end{enumerate}
Thus, depending on the condition met, the final path $P(x,y)$ is constructed as follows:
\begin{eqnarray}
P(x,y) = \{ (x^f_1,y^f_1) \dots (x^f_m,y^f_m) \}, \\
\nonumber \textrm{if condition (1) is met};\\
P(x,y) = \{ (x^b_n,y^b_n) \dots (x^b_1,y^b_1) \}, \\
\nonumber \textrm{if condition (2) is met};\\
P(x,y) = \{ (x^f_1,y^f_1) \dots (x^f_t,y^f_t), \\
\nonumber (x^b_{k-1},y^b_{k-1}) \dots (x^b_1,y^b_1) \}, \, \textrm{or} \\
\nonumber \{ (x^f_1,y^f_1) \dots (x^f_{k-1},y^f_{k-1}), (x^b_t,y^b_t) \dots (x^b_{1},y^b_{1}) \}, \\
\nonumber \textrm{if condition (3) is met}.
\end{eqnarray}
Here, $m$ and $n$ are the lengths of the forward and the backward path, respectively. In real applications, the points of the path $P(x,y)$ can then be used as via points to generate trajectories using conventional methods such as splines \cite{Egerstedt2001,Siciliano2009} or more advanced state-of-the-art methods such as dynamic movement primitives (DMPs, \cite{Ijspeert2013}), Gaussian mixture models (GMMs, \cite{Khansari2011}), probabilistic movement primitives (PMPs, \cite{Paraschos2013}) or optimal control primitives (OCPs, \cite{Herzog2017}). Note that in some cases paths could not be reconstructed, i.e., non-of the stopping conditions were met. In this case we treated network's prediction as failed (path not found).

\subsection{Evaluation measures and procedure}

We compared our approach against A* algorithm \cite{Hart1968}. We have chosen A* against rapidly-exploring random trees (RRT \cite{LaValle1998} or RRT* \cite{Karaman2011}), since it has been shown that algorithms such as Dijsktra and A* perform better on grid structures as compared to RRTs \cite{Knispel2013,Bency2019}.

Thus, for comparison we used the following criteria: 1) success rate, 2) path optimality and 3) run-time of the algorithm. Below we describe our evaluation measures and methods in more detail.

\textbf{Success rate:} We counted the network's prediction as successful (path found) if the path $P(x,y)$ could be reconstructed (see section \ref{sec:path_rec}) from the predicted output map $\hat{O}$, otherwise we counted network's prediction as failed (path not found). The success rate $SR$ was simply computed as percentage of successfully found paths out of all tested environments:
\begin{equation}
SR = 100\% \cdot \frac{N_S}{N_T},
\end{equation}
where $N_S$ is the number of  successfully predicted (found) paths and the $N_T$ is the total number of predicted paths.

\textbf{Path optimality:}
We also checked whether successfully predicted paths were optimal (i.e., shortest path) or not. For that, we compared path lengths of the paths obtained by using the FCN and A* algorithms, denoted as $L_{FCN}$ and $L_{A*}$, respectively. The path length was computed as follows (we skip $FCN$ and $A*$ notation for clarity):
\begin{equation}
L = \sum_{t=1}^{n-1} \|P(x_{t+1},y_{t+1})-P(x_{t},y_{t})\|,
\end{equation}
where $P(x,y)$ is the path predicted by the FCN or A* algorithm, and $n$ is the number of points in the path sequence. Here, $\|\cdot\|$ is the Euclidean norm. The path predicted by the FCN was counted as optimal if $L_{FCN} = L_{A*}$. Consequently, the percentage of optimal paths $OP$ was computed as
\begin{equation}
OP = 100\% \cdot \frac{N_O}{N_T},
\end{equation}
where $N_O$ is the number of optimal paths. The paths predicted by FCN are not always optimal, thus, we also computed the path length ratio $LR$ of non-optimal paths to analyse how much longer non-optimal paths are compared to the A* solutions:
\begin{equation}
LR = \frac{L_{FCN}}{L_{A^*}}.
\end{equation}

\textbf{Algorithm run-time:}
For run-time comparison we analysed how much time it takes to predict (in case of FCN) or to find (in case of A*) a single path for different path lengths. The path length was measured in steps, i.e., how many steps it takes to move from the start-point to the end-point. We expected on average longer run-times for longer paths in case of the A* algorithm and on average constant run-time in case of FCN. Note that for run-time comparison we excluded path reconstruction time for both A*\footnote{A* algorithm consists of two steps: path search and path reconstruction from visited states.} as well as FCN.

\textbf{Evaluation procedure:}
We performed two types of experiments: 1) prediction of single paths (one source and one target) and 2) prediction of multiple paths (two/three sources and one target). In the first case we were interested in the general performance of the system with respect to the above introduced measures and we wanted to check how well the network can generalise to different grid sizes. For this, we trained the network to predict single paths on three different grids ($10 \times 10$, $15 \times 15$ and $20 \times 20$) and then tested each model on all three grids.

In the second type of experiments we tested whether the network can predict multiple paths although the network has only been trained on single paths. In this study we checked the network's ability to predict two or three paths. Here we tested path predictions on the mid grid size, i.e., $15 \times 15$ (thus, the network model trained on $15 \times 15$ grid for single paths was used for testing).

\section{Results}

\subsection{General performance of the proposed network}

\begin{figure}[htb]
\begin{center}
\includegraphics[width=0.99\linewidth]{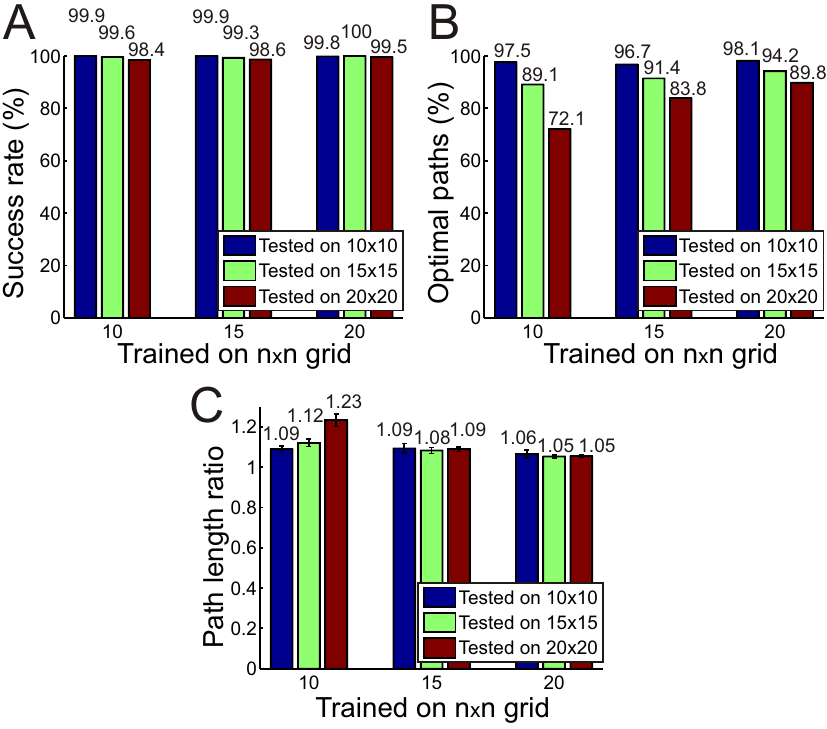}
\caption{\label{res_single} Results for the prediction of single paths obtained from 2000 tested samples on unknown environments. \textbf{A)} Success rate $SR$ of the predicted valid paths, \textbf{B)} percentage of the optimal paths $OP$ within successful predictions, and \textbf{C)} average path length ratio $LR$ of non-optimal paths (error-bars denote confidence intervals of mean [95\%]).}
\end{center}
\end{figure}

Examples of single path predictions for different grid sizes are shown in Fig.~\ref{exampl_single}, where the network prediction is marked by blue dots. The size of the dots corresponds to the network's output, where small dots have values close to zero and large dots have values close to one. The optimal solution, i.e. obtained using A* algorithm, is marked by crosses. In most of the cases (see statistical analysis in Fig.~\ref{res_single}) the network is able to predict optimal paths. Optimal paths, which correspond to the A* solution are shown in the first column, whereas in the second column we show examples of optimal paths which differ from the A* path (usually going around an obstacle from the other side). In the third and fourth column we show valid paths, which are suboptimal, i.e., longer (in most of the cases less than 10\%) than those found by A*. The last two columns show cases of failed predictions (paths which could not be reconstructed from the network's output). Note that in case of $10 \times 10$ grid we obtained only one case of a not found path out of 2000 tested environments.

\begin{figure}[htb]
\begin{center}
\includegraphics[width=0.85\linewidth]{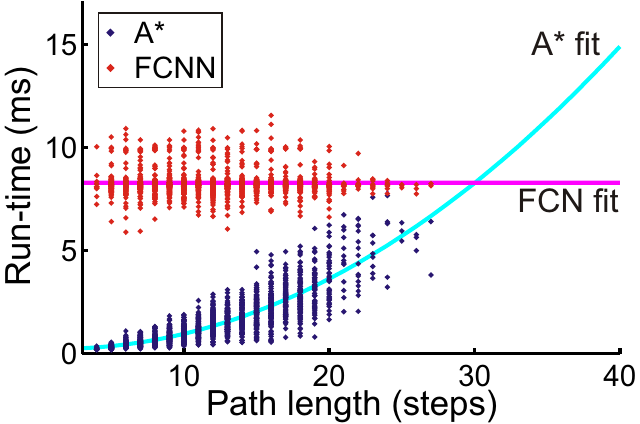}
\caption{\label{res_time} Results for the run-time comparison between A* and FCN obtained from 2000 tested samples on $20 \times 20$ grid. We used a first order and a second order polynomial functions to fit FCN and A* data, respectively.}
\end{center}
\end{figure}

A statistical evaluation of single path predictions is presented in Fig.~\ref{res_single}, where we show the network performance when trained and tested on different grid sizes. In general, we obtained a relatively high success rate, above 98\% (see panel A) in all cases, showing that the network can also predict paths quite reliably also on a grid size which was not used for training. As expected, we observe slightly worse performance of the model trained on $10 \times 10$ grid (99.9\%, 99.6\% and 98.4\% for $10 \times 10$, $15 \times 15$ and $20 \times 20$ grid, respectively) as compared to the model trained on $20 \times 20$ (99.8\%, 100\% and 98.5\%), however this difference is not statistically significant.

Although the network is able to predict a valid path (which leads from the start- to the end-point) in most of the cases, it is not always optimal. The number of optimal paths, within successful predictions, is shown in Fig.~\ref{res_single}~B, where we observe that we obtain less optimal paths in larger environments (from 72.1\% to 89.8\% for the test case $20 \times 20$) as compared to  smaller environments (from 96.7\% to 98.1\% for the test case $10 \times 10$). This is due to the fact that paths in larger environments on average are longer than in smaller environments, and thus are more prone to prediction errors. We can also see that the number of optimal paths increases if larger environments are used for training, since smaller environments do not include longer paths in the training set.

Next we checked how much longer non-optimal paths are with respect to the shortest path. The results are shown in Fig.~\ref{res_single}~C. As in the previous cases, a better performance is obtained if trained on the same or larger size grids than the test case. Nevertheless, except for the two test cases ($15 \times 15$ and $20 \times 20$ when trained on the $10 \times 10$), the non-optimal paths on average are below 10\% longer compared to the shortest path (in the range between 5\% and 9\%).

\begin{figure*}[htb]
\begin{center}
\includegraphics[width=0.99\linewidth]{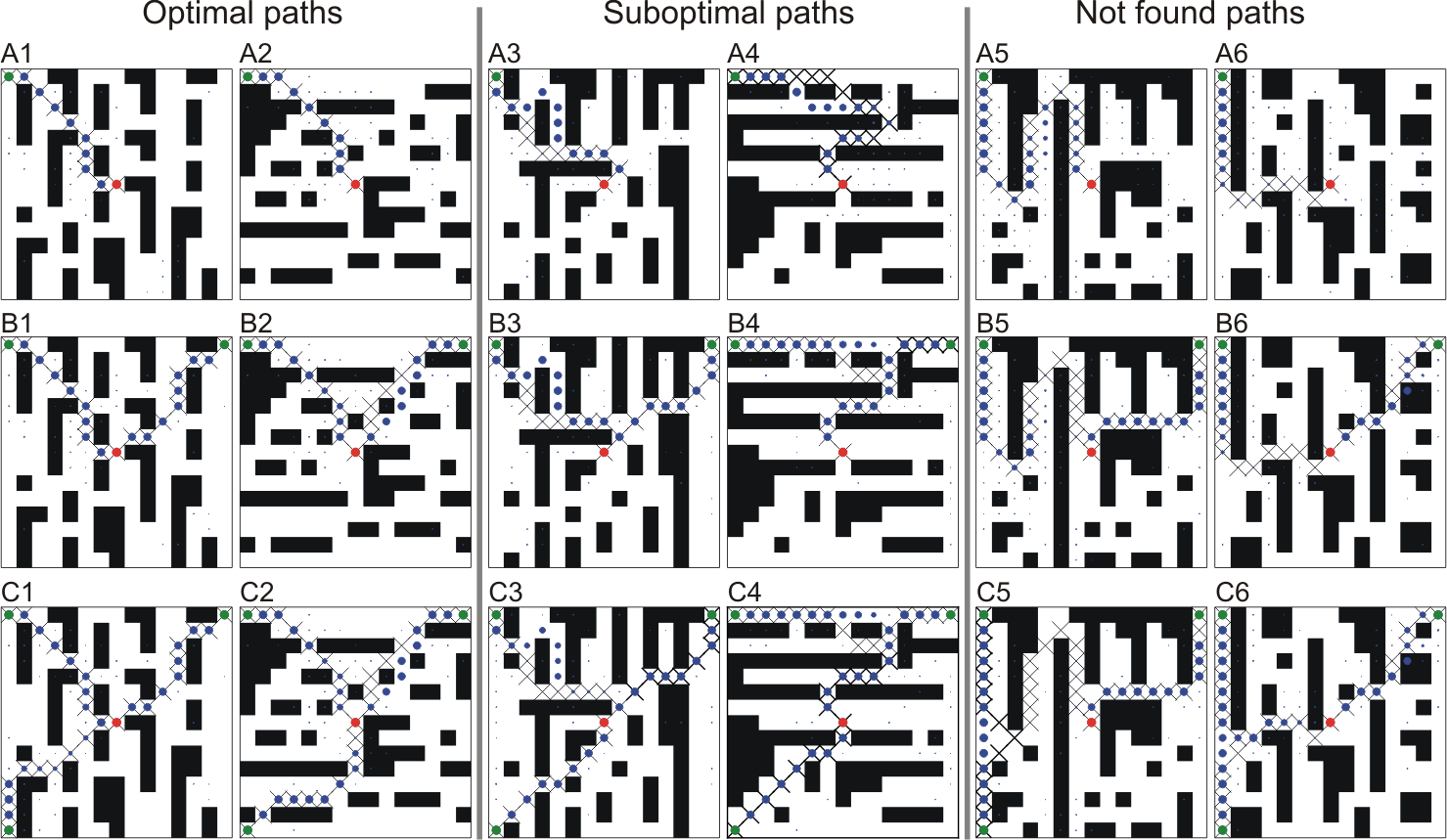}
\caption{\label{exampl_multi} Examples of multi-path predictions on unknown environments for grid size $15 \times 15$. \textbf{A)} Single path prediction (control case), \textbf{B)} predictions of two paths, and \textbf{C)} prediction of three paths. The first two columns show optimal (shortest) paths, the middle two columns suboptimal paths and the last two columns not found paths. Crosses denote A* solutions where blue dots denote predicted paths using FCN. Size of the dots correspond to  small (close to zero) and large (close to one) values of the network output. Green and red dots correspond to start-points and end-point, respectively. Note that the predicted path shown in panel A4 is optimal, and that the predicted path shown in panel A5 is a valid path (was found).}
\end{center}
\end{figure*}

In the last experiment we compared run-time, i.e., how much time it takes to process one environment, between A* and FCN. Results obtained on the $20 \time 20$ grid are presented in Fig.~\ref{res_time}. Here we show run-time of each realization (in total 2000 samples) for both A* and FCN. As expected, results show that on average run-time of A* increases quadratically if paths are getting longer, whereas run-time of FCN on average stays constant. Fitted functions predict that for paths longer than 30 steps (e.g., in larger environments), FCN would outperform A*. Also, as we show next, our proposed network is able to predict multiple paths within the same time, whereas in case of A* one would need to run path search several times, which --- as a consequence --- would scale run-time by $k$, where $k$ is the number of searched paths.

\subsection{Prediction of multiple paths}

Examples for predictions of two and three paths are shown in Fig.~\ref{exampl_multi}. Note that the first row is a control case. As above, also here we show cases of optimal, suboptimal and not found paths. The following four cases can be observed.
\begin{enumerate}
\item
Paths are not disturbed by adding a second or third source (usually leads to optimal solutions, see optimal paths).

\item
A path can become suboptimal when adding more sources. For example, the path shown in panel A4 (optimal) becomes suboptimal when adding the second source (see panel B4).

\item
A path can ``disappear'' when adding a second or third source (see panels A5, B5 and C5), or the other way around, 

\item
A path can ``appear'' when adding more sources (see panels A6 and C6).
\end{enumerate}

\begin{figure}[htb]
\begin{center}
\includegraphics[width=0.85\linewidth]{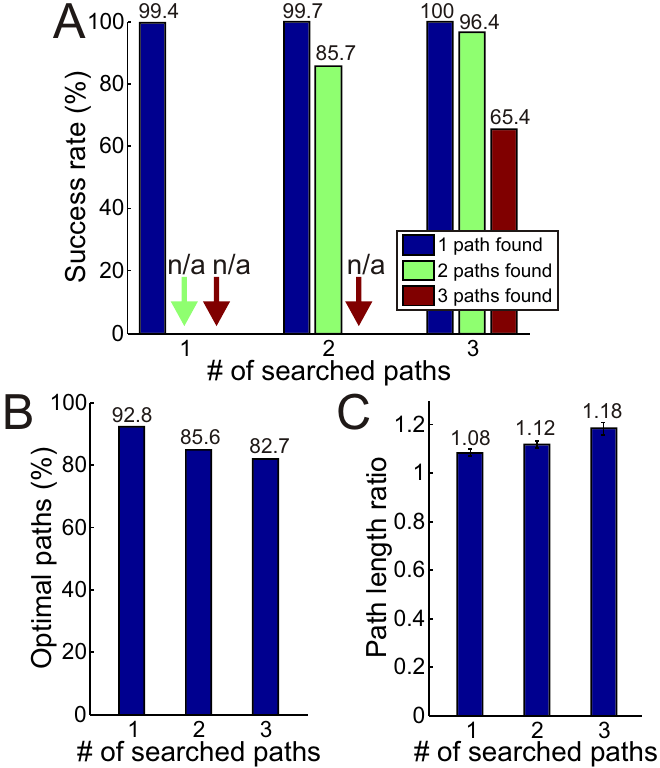}
\caption{\label{res_multi} Results for the prediction of multiple paths obtained from 1000 tested samples on unknown environments. \textbf{ A)} Success rate $SR$ of the predicted valid paths, \textbf{B)} percentage of optimal paths $OP$ within successful predictions, and \textbf{C)} average path length ratio $LR$ of non-optimal paths (error bars denote confidence intervals of mean [95\%]).}
\end{center}
\end{figure}

A statistical evaluation for this is given in Fig.~\ref{exampl_multi} where we show the performance for prediction of one (control case), two and three paths. Results show that the success rate of finding two paths out of two searched paths and three paths out of three searched paths is 85.7\% and 65.7\%. Thus, performance is decreasing with increased number of searched paths. On the other hand, in the case of three-path-search the network was always able to predict at least one path, and success rate for the prediction of two paths was relatively high 96.4\% (we get an improvement compared to two-path-search since there is a higher chance to predict two paths out of three than two out of two).

The analysis of path optimality is presented in Fig.~\ref{exampl_multi}~B and C. We observe that we get fewer optimal paths (85.6\% and 82.7\% for two-path and three-path search, respectively) and that paths become longer (12\% and 18\% for two-path and three-path search, respectively) with increased number of searched paths. This is due to the fact that in the case of multiple sources, the paths are more prone to intersect and this way become less optimal. However, given the fact that the network has never been trained on multiple paths, it shows a surprisingly good performance. Naturally, results should improve by training the network on multiple paths, but the goal of this study was to show how far one can even get by not doing this.

\section{Discussion}
In this work, we have presented a novel approach for the generation of single as well as multiple paths. To the best of our knowledge, this is the first approach, which allows predicting complete multiple paths while running the network's prediction only once. Note that most of the afore discussed deep-learning approaches generate paths iteratively and only deal with a planning of single paths \cite{Tai2017,Panov2018,Qureshi2018,Bency2019} or generate multiple-paths for each agent separately \cite{Long2017,Chen2017,Everett2018} but only deal with collision avoidance path planning and not with path planning of paths for navigation in maze-like environments.

Recently, \cite{Perez2018} proposed an approach, which is also able to plan paths in a one-shot using fully convolutional network, however, they only deal with single path planning for human-aware collision-free navigation in simple environments. Also, they use a two-step approach where first the network is used to predict the path and then RRT* \cite{Karaman2011} is used on top of it to refine the predicted path, which is computationally more expensive than the here used simple path reconstruction algorithm (bidirectional search).

We have demonstrated that in case of single path predictions the proposed network is able to predict optimal or close to optimal paths successfully in more than 99\% of the cases (when trained and tested on the same grid size). Moreover, we have also shown that our proposed the network can be trained on one size grid and then used on a different size grids (success rate is above 98\% in all tested cases). 

In case of multiple path predictions, we have shown that, although the network has never been trained on multiple paths, it is also able to generate paths from multiple sources to one target. This could be also used to solve single-source-multi-targets problems. We obtained a success rate of 85.7\% and 65.4\% for the prediction of two and three paths, respectively. As mentioned above, we believe that the performance can be improved by including samples of multiple paths in the training set. The other option would be to repeat the network's prediction one or several more times by giving locations at which the path reconstruction was lost as new start- and end-points.

In our study we have only dealt with static environments. However, our approach can also be used for on-line planning/replanning in dynamic environments. In this case, the current position of the robot can be used as the start-point and the next sub-goal or the final goal as the end-point.

\section*{Acknowledgements}
The research leading to these results has received funding from the  European Community’s H2020 Programme (Future and Emerging Technologies,
FET) under grant agreement no. 732266, Plan4Act.

\bibliographystyle{plain}


\end{document}